\documentclass[10pt,twocolumn,letterpaper]{article}

\usepackage[pagenumbers]{cvpr} 

\usepackage{graphicx}
\usepackage{amsmath}
\usepackage{amssymb}
\usepackage{booktabs}
\usepackage{gensymb}
\usepackage[numbers,sort]{natbib}
\usepackage{multirow}
\usepackage{graphicx}
\usepackage{makecell}

\usepackage{color}
\usepackage[numbers,sort]{natbib}
\usepackage{multirow}
\usepackage{array}

\usepackage{pifont}
\definecolor{hollywoodcerise}{rgb}{0.96, 0.0, 0.63}
\definecolor{lasallegreen}{rgb}{0.03, 0.47, 0.19}
\definecolor{hanpurple}{rgb}{0.32, 0.09, 0.98}
\definecolor{green(pigment)}{rgb}{0.0, 0.65, 0.31}

\usepackage[pagebackref=true,breaklinks=true,letterpaper=true,colorlinks,bookmarks=false]{hyperref}

\hypersetup{colorlinks,linkcolor={red},citecolor={hollywoodcerise},urlcolor={red}}  

\usepackage[capitalize]{cleveref}
\crefname{section}{Sec.}{Secs.}
\Crefname{section}{Section}{Sections}
\Crefname{table}{Table}{Tables}
\crefname{table}{Tab.}{Tabs.}

\begin{document}

\title{BIPS: Bi-modal Indoor Panorama Synthesis via Residual Depth-aided Adversarial Learning}

\author{Changgyoon Oh\thanks{The first two authors contributed equally. In alphabetical order. }\;, Wonjune Cho$^{*}$, Daehee Park, Yujeong Chae, Lin Wang and Kuk-Jin Yoon\\
Visual Intelligence Lab., KAIST, Korea\\
{\tt\small \{changgyoon,wonjune,bag2824,yujeong,wanglin,kjyoon\}@kaist.ac.kr}}

\maketitle

\begin{abstract}

Providing omnidirectional depth along with RGB information is important for numerous applications, \eg, VR/AR. However, as omnidirectional RGB-D data is not always available, synthesizing RGB-D panorama data from limited information of a scene can be useful. Therefore, some prior works tried to synthesize RGB panorama images from perspective RGB images; however, they suffer from limited image quality and can not be directly extended for RGB-D panorama synthesis. In this paper, we study a new problem: RGB-D panorama synthesis under the arbitrary configurations of cameras and depth sensors. Accordingly, we propose a novel bi-modal (RGB-D) panorama synthesis (BIPS) framework. Especially, we focus on indoor environments where the RGB-D panorama can provide a complete 3D model for many applications. We design a generator that fuses the bi-modal information and train it with residual-aided adversarial learning (RDAL). RDAL allows to synthesize realistic indoor layout structures and interiors by jointly inferring RGB panorama, layout depth, and residual depth. In addition, as there is no tailored evaluation metric for RGB-D panorama synthesis, we propose a novel metric to effectively evaluate its perceptual quality. Extensive experiments show that our method synthesizes high-quality indoor RGB-D panoramas and provides realistic 3D indoor models than prior methods. Code will be released upon acceptance.

\end{abstract}

\vspace{-14pt}
\section{Introduction}
\label{sec:intro}
\vspace{-5pt}

\begin{figure}[t]
\begin{center}
  \includegraphics[width=\linewidth]{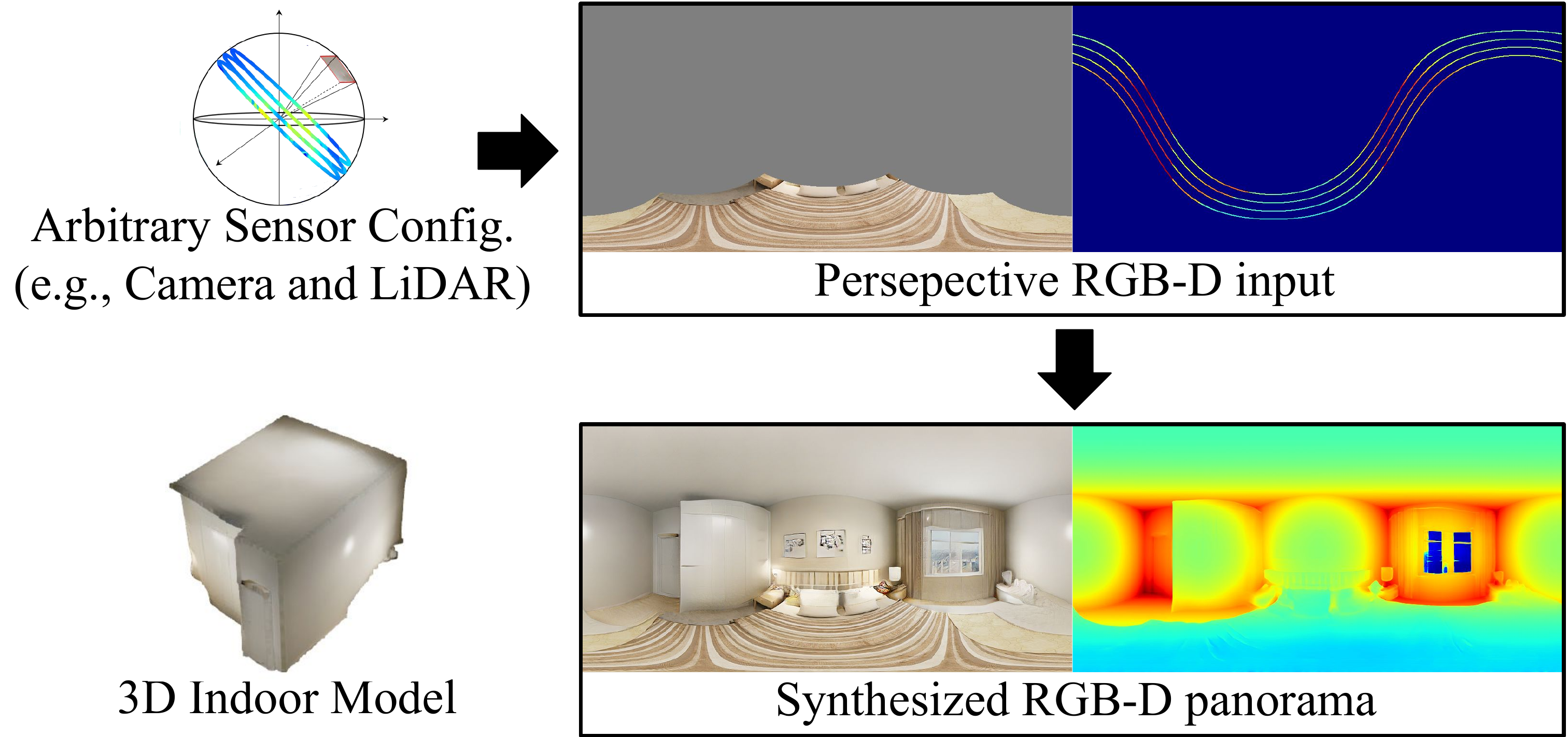}
\end{center}
\vspace{-13pt}
  \caption{Overall scheme of RGB-D panorama synthesis. Our method takes RGB-D input from cameras and depth sensors in arbitrary configurations and synthesizes an RGB-D panorama.}
\label{fig:intro}
\vspace{-18pt}
\end{figure}

Providing omnidirectional depth along with RGB information is important for numerous applications, \eg, VR/AR. 
However, as the omnidirectional RGB-D data is not always available, synthesizing RGB-D panorama data from the limited information of the scene can be useful.
Even though prior works have  tried to
 synthesize RGB panorama images from perspective RGB images~\cite{hara2020spherical,sumantri2020360},
these methods show limited performance on synthesizing panoramas from small partial views and can not be directly extended for RGB-D panorama synthesis.




By contrast, \textit{jointly learning to synthesize depth data along with the RGB images} allows to synthesize RGB-D panorama with two distinct advantages: (1) RGB images and depth data share the semantic correspondence that can improve the quality of the output RGB-D panorama. (2) Synthesized depth panorama provides omnidirectional 3D information, which can be potentially applied to plentiful applications. 
Therefore, it is promising to synthesize RGB-D panorama from the cameras and depth sensors, such that we can synthesizing realistic 3D indoor models.  

\begin{figure*}[t!]
\begin{center}
   \hspace*{-0.0\linewidth}\includegraphics[width=.94\textwidth]{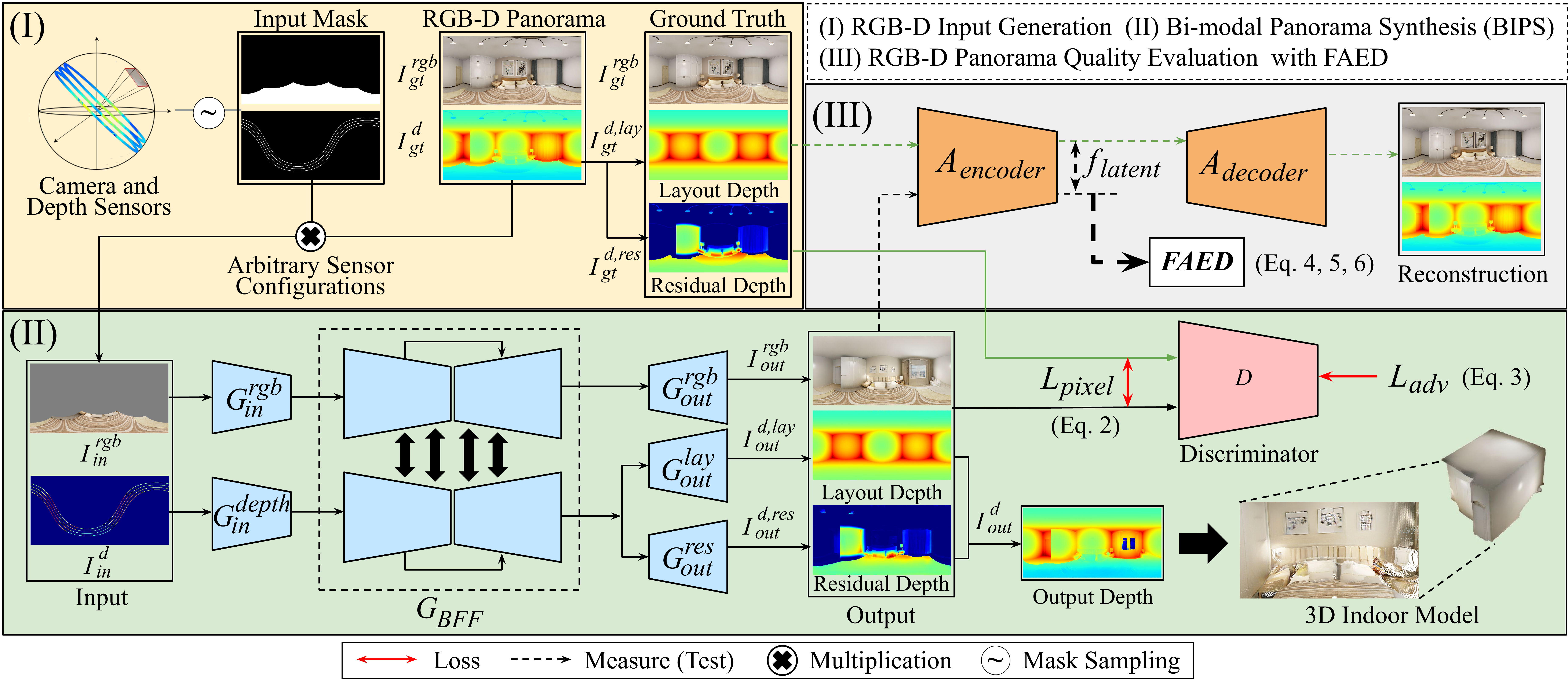} 
\end{center}
   \vspace{-0.5cm}
   \caption{Overall structure of our bi-modal indoor panorama  synthesis (BIPS) framework. Our framework takes RGB-D input provided by arbitrary sensor configurations, integrates the bi-modal input data with BFF branch in generator network, and jointly trains to synthesize layout depth and residual depth. Then, perceptual quality of the synthesized RGB-D panorama is measured by our proposed FAED metric.}
   \vspace{-14pt}
\label{fig:overall}
\end{figure*}

In this paper, we consider a novel problem: \emph{RGB-D panorama synthesis from limited input information about a scene}. To maximize the usability, we consider the arbitrary configurations of cameras and depth sensors.
To this end, we design the arbitrary sensor configurations by randomly sampling the number of sensors, their intrinsic parameters, and extrinsic parameters, assuming that the sensors are calibrated such that we can align the depth data with the RGB image. 
This enables to represent most of the possible combinations of cameras and depth sensors. Accordingly, we propose a novel bi-modal panorama synthesis (BIPS) framework to synthesize RGB-D indoor panoramas from the camera and depth sensors in arbitrary configurations via adversarial learning (See Fig.~\ref{fig:overall}). Especially, we focus on the indoor environments as the RGB-D panorama can provide the complete 3D model for many applications.
We thus design a generator that fuses the bi-modal (RGB and depth) features. Through the generator, multiple latent features from one branch can help the other by providing the relevant information of different modality. 

For synthesizing the depth of \emph{indoor} scenes, we rely on the fact that the overall layout usually made of flat surfaces, while interior components have various structures. 
Thus, we propose to separate the depth of a scene $I^{d}$ into two components: layout depth $I^{d,lay}$ and residual depth $I^{d,res}$. Here, $I^{d,lay}$ corresponds to the depth of planar surfaces, and $I^{d,res}$ corresponds to the depth of other objects, \eg, furniture. 
With this relation, we propose a joint learning scheme called \textit{Residual Depth-aided Adversarial Learning (RDAL)}. 
RDAL jointly trains RGB panorama, layout depth and residual depth to synthesize more realistic RGB-D panoramas and 3D indoor models (Sec.~\ref{generator}). 

Previously, some metrics \cite{salimans2016improved, heusel2017gans} have been proposed to evaluate the outputs of generative models using latent feature distribution of a pre-trained classification network \cite{szegedy2016rethinking}. However, the input modality of utilizing off-the-shelf network is only limited to perspective RGB images. 
Therefore, a new tailored evaluation metric for RGB-D panoramas is needed. For this reason, we propose a novel metric, called Fréchet Auto-Encoder Distance (FAED),
to evaluate the perceptual quality for RGB-D panorama synthesis (Sec.~\ref{subsec:FAED}).
FAED adopts an auto-encoder to reconstruct the inputs from latent features with unlabeled dataset. 
Then, the latent feature distribution of the trained auto-encoder is used to calculate the Fréchet distance between the synthesized and real RGB-D data. 

Extensive experimental results demonstrate that our RGB-D panorama synthesis method significantly outperforms 
the extensions of the prior image inpainting \cite{marinescu2020bayesian, zhao2021large, suvorov2021resolution}, image outpainting \cite{boundless, sumantri2020360}, and image-guided depth synthesis methods \cite{Cheng_2020_TPAMI, park2020non, Li_2020_WACV, hu2020PENet} modified to synthesize RGB-D panorama from partial arbitrary RGB-D inputs. 
Moreover, we show the validity of the proposed FAED for evaluating the quality of synthesized RGB-D panorama by showing how well it captures the disturbance level \cite{heusel2017gans}.

In summary, our main contributions are three-fold: (I) We introduce a new problem of generating RGB-D panoramas from partial and arbitrary RGB-D inputs. 
(II) We propose a BIPS framework that allows to synthesize RGB-D panoramas via residual depth-aided adversarial learning. 
(III) We introduce a novel evaluation metric, FAED, for RGB-D panorama synthesis and demonstrate its validity.



\section{Related Works}
\vspace{-5pt}
\label{sec:relatedworks}

\noindent \textbf{Image Inpainting}
Conventional approaches explore diffusion or patch matching~\cite{ballester2001filling, barnes2009patchmatch, bertalmio2000image, criminisi2003object, efros1999texture, bertalmio2003simultaneous, criminisi2004region}. 
However, they require visible regions sufficiently enough to inpaint the missing regions, 
thus limiting their ability to synthesize novel textures or structures. The learning-based methods often use generative adversarial networks (GANs) to synthesize texture or structures~\cite{zheng2019pluralistic, li2017generative, iizuka2017globally, yu2018generative}, optimized by the minimax loss~\cite{isola2017image}. 
Some works explored different convolution layers, \eg, partial convolution~\cite{liu2018image} and gated convolution~\cite{yu2019free,navasardyan2020image}, to better handle invalid pixels in the input data to the convolution kernel. 
Moreover, attention mechanism~\cite{vaswani2017attention, wan2021high} has also been applied to better capture the contextual information and handle missing contents~\cite{yu2018generative, xie2019image, liu2019coherent, wang2019musical, liao2021image}.
Recently, research has been made to synthesize high-resolution outputs~\cite{suin2021distillation, wang2021parallel, peng2021generating} or semantically diverse outputs~\cite{liu2021pd, zhao2020uctgan}. 
Although endeavours have been made to tackle this large completion problem \cite{marinescu2020bayesian, zhao2021large, suvorov2021resolution}, they often fail to synthesize visually pleasing panoramas due to only using perspective RGB inputs. 


\noindent\textbf{Image Outpainting} 
Conventional methods extend an input image to a larger seamless one; however, they require manual guidance \cite{seamcarving, patchmatch, zhang2013framebreak} or image sets of the same scene category \cite{infiniteimages, photouncrop, biggerpicture}.
By contrast, learning-based methods synthesize large images with novel textures that do not exist in the input perspective image~\cite{sabini2018painting, unsup_holistic, deep_portrait, structureaware, 9191339, multimodalout, Mastan2021DeepCFLDC, spiral, Kim2021PaintingOA}.
Some approaches focus on driving scenes \cite{widecontext, sienet} or synthesize panorama-like landscapes with iterative extension or multiple perspective images \cite{yang2019very, boundless,hara2020spherical,sumantri2020360}. 
Although performance has been greatly improved so far, the existing methods are still afflicted by the limited quality from the perspective images. 

\noindent \textbf{Image-guided Depth Synthesis}
One line of research attempts to fuse the bi-modal information,  \ie, the RGB image and sparse depth. Some methods, \eg~\cite{mal2018sparse}, fuse the sparse depth and RGB image via early fusion while others~ \cite{self_sparsetodense, jaritz2018sparse, tang2019learning, 2020, li2020multi, Huang_2020_TIP} utilize a late fusion scheme,
or jointly utilize both the early and late fusion~\cite{Gansbeke_2019_MVA, Lee_2020_Access, wang2020fisnets}. Another line of research focuses on utilizing affinity or geometric information of the scene via surface normal, occlusion boundaries, and the geometric convolutional layer~\cite{Lee_2019_ICRA, Qiu_2019_CVPR, Xu_2019_ICCV, Zhang_2018_CVPR, hu2020PENet,Cheng_2020_TPAMI,cspn++, park2020non}. 
However, these works only generate dense depth maps that has the same FoV as the input perspective RGB images. 

\noindent \textbf{Evaluation of Generative Models}
Image quality assessment can be classified into three groups: full-reference (FR), reduced-reference (RR), and no-reference (NR). 
There exist many conventional FR metrics, \eg, PSNR, MSE, and SSIM, and deep learning (DL)-based FR metrics, \eg, LPIPS \cite{zhang2018unreasonable}. These metrics typically calculate either pixel-wise, or patch-wise similarity to the ground truth images. 
By contrast, NR methods, \eg, BRISQUE~\cite{mittal2012no} and NIQE~\cite{mittal2012making}  assess image quality without any reference image.  
Among the DL-based NR metrics, Inception Score (IS) ~\cite{salimans2016improved} and Fréchet Inception Distance (FID)~\cite{heusel2017gans} are two popular approaches~\cite{borji2019pros}. 
IS and FID scores are calculated based on pretrained classification models, \eg, Inception model~\cite{szegedy2016rethinking}, aiming to capture the high-level features. 
Unfortunately, these metrics are less applicable for RGB-D panorama evaluation because (1) they are trained only with perspective RGB images, and (2) there are no labeled panorama images to train them.
Therefore, they are highly sensitive to the distortion of panoramic images,  making them hard to capture perceptual quality properly on panoramic images.
Furthermore, naively using them on RGB-D information leads to imprecise measure of the semantic correspondence between the two different modalities. Therefore, \textit{we propose FAED, which aims to evaluate RGB-D panorama quality between the RGB and depth pairs. 
FAED can be adaptively applied to generative models on multi-modal domain, that lacks labeled dataset}.


\vspace{-5pt}
\section{Proposed Methods}
\vspace{-2pt}

\vspace{-2pt}
\subsection{Problem Formulation}
\vspace{-3pt}
\label{subsubsec:simulation}
Previous works, \eg, \cite{sumantri2020360, hara2020spherical} generate an equirectangular projection (ERP) image ($ERP^{rgb}$) from input perspective image(s) ($I^{rgb}_{in}$).
Then, an RGB panorama $I^{rgb}_{out}$ can be created via a function $G$, mapping $I^{rgb}_{in}$ into a $ ERP^{rgb}$~\cite{he2017geometry}, which can be formulated as $I^{rgb}_{out} = ERP^{rgb}=G(I^{rgb}_{in})$.

However, as it is crucial to provide omnidirectional depth information \cite{rosin2019rgb, alaee2018user} in many applications, many studies tried to synthesize depth panoramas from input RGB panorama images and partial depth measurements~\cite{wang2020bifuse, hirose2021depth360}.
One solution to synthesize an RGB-D panorama would be to first synthesize RGB panorama from input perspective images, and then utilizing the depth synthesis methods to generate an omnidirectional depth map. However, such an approach is cumbersome and less effective, as shown in the experimental results (See Table~\ref{tab:ablation}). 
We solve this novel yet challenging problem by jointly utilizing the input RGB image ($I^{rgb}_{in}$) and depth data ($I^{d}_{in}$). Our goal is to directly generate the RGB panorama ($ERP^{rgb}$) and depth panorama ($ERP^{d}$) simultaneously via a mapping function $G$, which can be described as $(I^{rgb}_{out},I^{d}_{out})=(ERP^{rgb},ERP^{d})= G(I^{rgb}_{in}, I^{d}_{in})$.  
$G$ can be formulated by learning a \textit{single} network to synthesize $ERP^{rgb}$ and $ERP^{d}$ using 
$I^{rgb}_{in}$ and $I^{d}_{in}$ obtained in arbitrary sensor configurations.
As the information in the left and right boundaries in ERP images should be connected, our designed G uses circular padding~\cite{schubert2019circular} before each convolutional operation.





\begin{figure}[t]
\begin{center}
   \hspace*{-0.0\linewidth}\includegraphics[width=.96\linewidth]{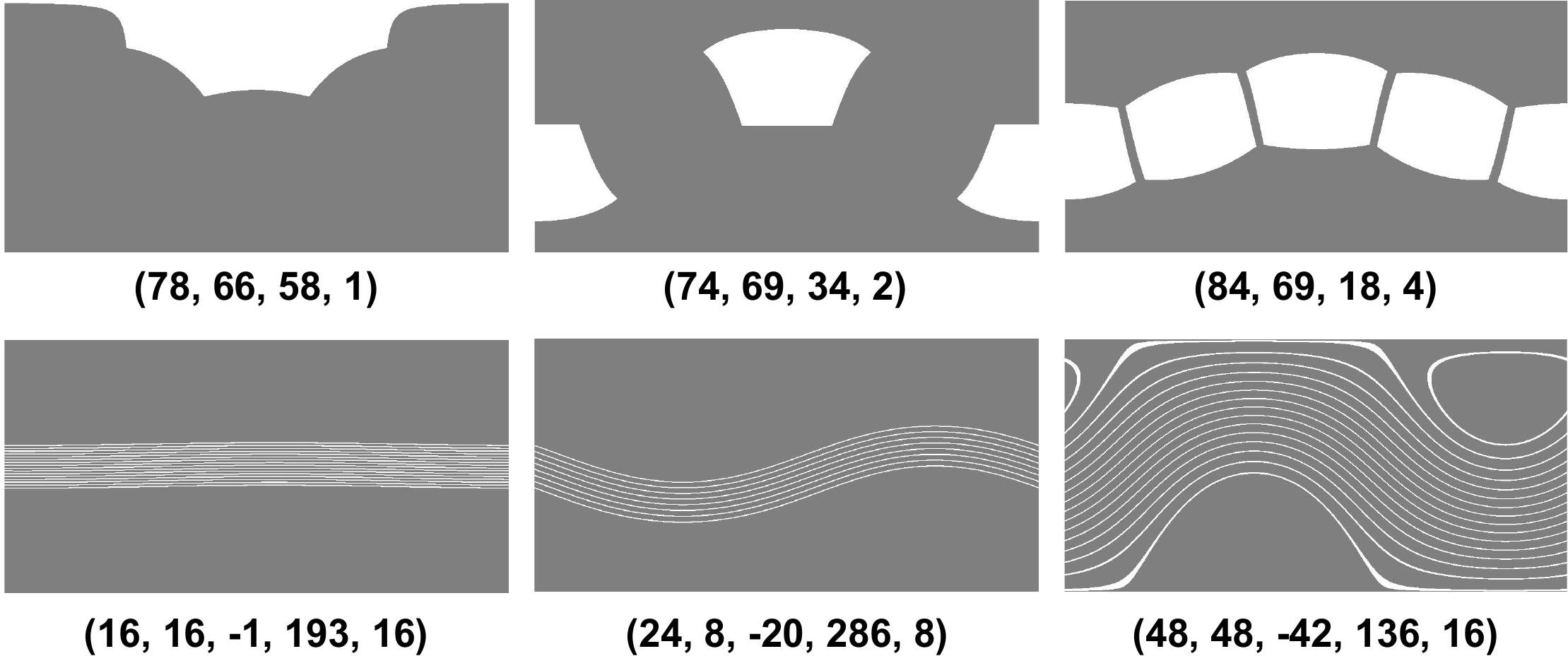}
\end{center}
   \vspace{-0.6cm}
   \caption{Sampled input masks. The upper row shows the visible regions of cameras and perspective dense depth sensors with parameters $(\delta_{H}, \delta_{V}, \psi, n)$, and the lower row shows the visible regions of mechanical LiDARs, with parameters $(\delta_{L}, \delta_{U}, \psi, \omega, \eta)$.}
\label{fig:input_mask} 
\vspace{-14pt}
\end{figure}

Consequently, we configure the parameters of cameras and depth sensors, and randomly sample the parameters to provide the input to the $G$ during training. 
These parameters can handle most of the possible sensor configurations. 
Figure~\ref{fig:input_mask} shows the input masks, sampled from the sensor configurations. To handle the cases where only cameras or depth sensors are used, we choose whether to use cameras only, depth sensors only, or both, randomly.

\noindent \textbf{Parameters of RGB Cameras} We denote the parameters of RGB cameras, horizontal FoV as $\delta_{H}$, vertical FoV as $\delta_{V}$, pitch angle as $\psi$, and number of viewpoints as $n$. When $n>1$, we arrange the viewpoints in a circle having the sampled pitch angle from the equator and at the same intervals.
We do not consider roll and yaw, as they 
do not affect the results (\ie, the output is equivariant to the horizontal shift of input) thanks to using circular padding. 
Practically, we sample the parameters from $\delta_{H} \sim \mathcal{U}[60^{\circ}, 90^{\circ}]$, $\delta_{V} \sim \mathcal{U}[60^{\circ}, 90^{\circ}]$, $\psi \sim \mathcal{U}[-90^{\circ}, 90^{\circ}]$, and $n \sim \mathcal{U}\{1,2,3,4\}$, where $\mathcal{U}\left [ \cdot  \right ]$ represents uniform distribution.

\noindent \textbf{Parameters of Depth Sensors} $I^{d}_{in}$ can be obtained from mechanical LiDARs or perspective depth sensors, thus we should generate arbitrary depth input masks for both. For the LiDARs, we denote the parameters as lower FoV $\delta_{L}$, upper FoV $\delta_{U}$, pitch angle $\psi$, yaw angle $\omega$, and the number of channels $\eta$. The yaw angle is needed here to consider the relative yaw motion to the camera arrangement. 
For the perspective depth sensors providing dense depth, they have many similar parameters and viewpoints with those of the RGB-D cameras. Therefore, we use the same sampled parameters with the cameras (\ie, $(\delta_{H}, \delta_{V}, \psi, n)$). 
In practice, we first sample the parameters from $\psi \sim \mathcal{U}[-90^{\circ}, 90^{\circ}]$, $\omega \sim \mathcal{U}[0,360^{\circ}]$, and $\eta \sim \mathcal{U}\{2, 4, 8, 16\}$. Then, we sample $\delta_{L}$ and $\delta_{U}$ from $\mathcal{U}\{\eta, 2\eta, 3\eta\}$.
Finally, our problem is formulated as:
\vspace{-5pt}
\begin{equation}
\begin{split}
(I^{rgb}_{out},I^{d}_{out})=(ERP^{rgb},ERP^{d})= \\ G(I^{rgb}_{in}(\delta_{H}, \delta_{V}, \psi, n), I^{d}_{in}(\delta_{L}, & \delta_{U}, \psi, \omega, \eta, \delta_{H}, \delta_{V}, n)) 
\end{split}
\end{equation}
\vspace{-5pt}

\vspace{-15pt}
\subsection{RGB-D Panorama Synthesis Framework}
\label{subsec:network_architecture}
\vspace{-0.1cm}

\noindent\textbf{Overview} An overview of the proposed BIPS framework is depicted in Fig.~\ref{fig:overall}. BIPS consists of a generator $G$ (Sec.\ref{generator}) and a discriminator $D$ (Sec.~\ref{disc}). $G$ takes the perspective RGB image $I^{rgb}_{in}$ and depth $I^d_{in}$ as inputs. 
We notice that the quality of the RGB-D panorama depends on both the overall (mostly rectangular) layout and how the furniture are arranged in the indoor scene. Inspired by \cite{zeng2020joint}, we separate the depth data $I^d_{gt}$ into \textit{layout depth} $I^{d,lay}_{gt}$, and \textit{residual depth (the interior components) $I^{d,res}_{gt}$} which is defined as ($I^d_{gt}$ - $I^{d,lay}_{gt}$ ). The generator $G$ outputs the RGB panorama image $I^{rgb}_{out}$, the layout depth $I^{d,lay}_{out}$ and residual depth $I^{d,res}_{out}$ simultaneously. As these are jointly trained with adversarial loss, we call this learning scheme as \textit{Residual Depth-aided Adversarial Learning (RDAL)}.

\vspace{-12pt}
\subsubsection{Generator}
\vspace{-5pt}
\label{generator}
\noindent \textbf{Input Branch} 
$G_{in}$ consists of two encoding branches, $G_{in}^{rgb}$ and $G_{in}^{depth}$ which takes $I^{rgb}_{in}$ and $I^d_{in}$, respectively. 
These branches independently process $G_{in}^{rgb}$ and $G_{in}^{depth}$ with six conv layers before fusing them. 
As the inputs have a resolution of 512 $\times$1024, the filter size of the first conv layer is set as 7 and then reduced to 3 and 4 gradually.

\begin{figure}[t]
\begin{center}
   \hspace*{-0.0\linewidth}\includegraphics[width=\linewidth]{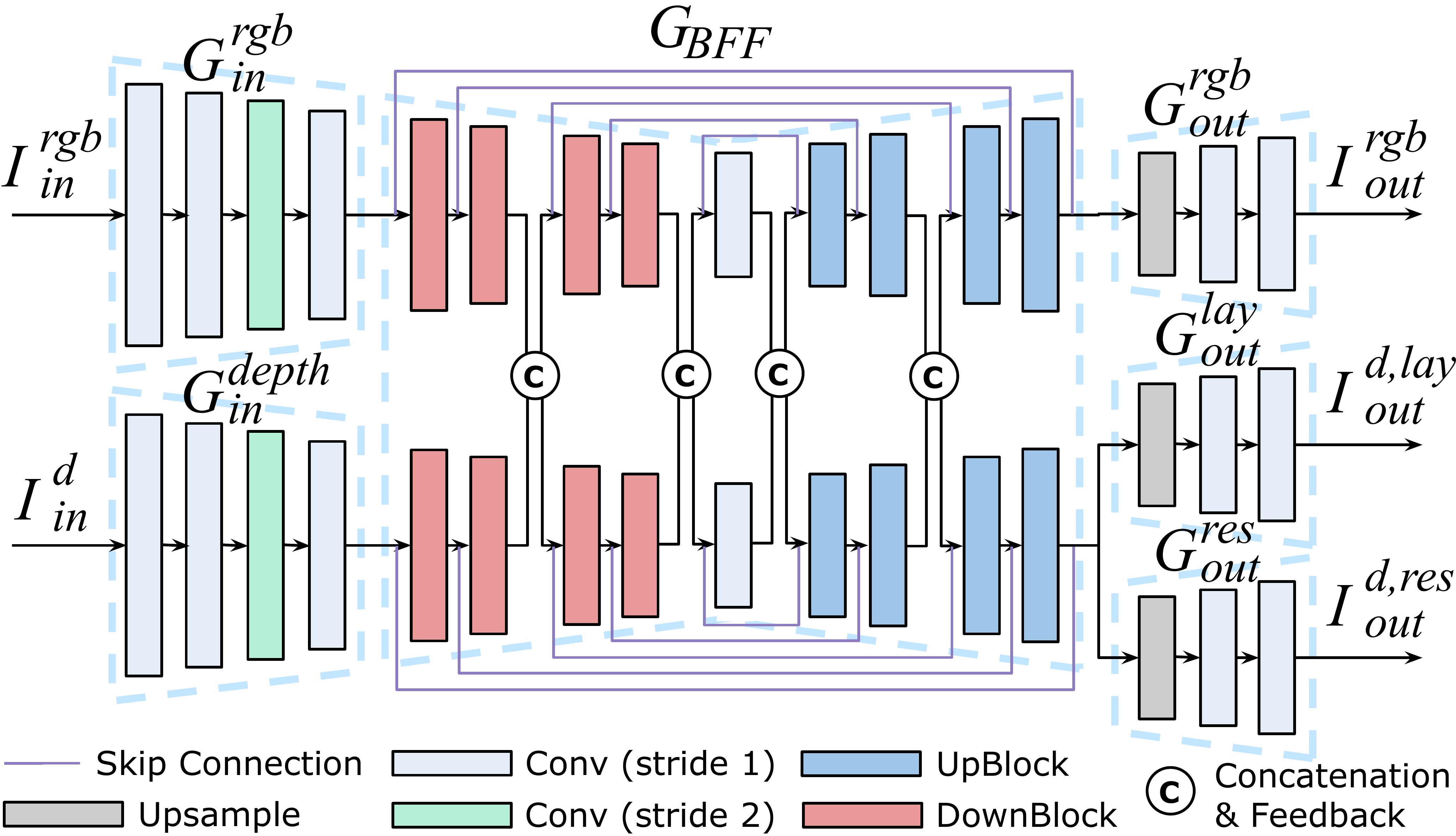}
\end{center}
   \vspace{-0.4cm}
   \caption{The proposed generator ($G$). It consists of two input branches, a BFF branch, and three output branches. A larger version of the image can be found in the suppl. material.}
   \vspace{-0.5cm}
\label{fig:generator}
\end{figure}

\noindent \textbf{Bi-modal Feature Fusion (BFF) Branch } 
 BFF branch $G_{BFF}$ takes $G_{in}^{rgb} (I^{rgb}_{in})$ and $G_{in}^{depth} (I^{d}_{in})$ as inputs, as shown in Fig.~\ref{fig:generator}. Although  $I^{rgb}_{in}$ and depth $I^d_{in}$ are in two different modalities, we assume that the cameras and depth sensors are well calibrated and synchronized. 
Then, to utilize this highly correlated bi-modal information in its two branches, $G_{BFF}$ consists of two-stream encoder-decoder networks fusing the bi-modal features. These two encoder-decoder networks have an identical structure (see Fig.~\ref{fig:generator}). 

Moreover, the bi-modal features are fused in between the layers of $G_{BFF}$. In particular, the features from both branches are concatenated and fed back to each other. Overall, the fusion is done after the features pass two `DownBlocks' and before passing two `UpBlocks'.  In addition, multi-scale residual connections are used to vitalize transfer of information between the layers and branches. As multiple latent features from one branch help the other by sharing the information apart in both ways, $G_{BFF}$ can generate features by fully exploiting the information of the 3D scene.

\noindent \textbf{The Output Branch} 
Realistic indoor space comes from precise layout structure 
and high perceptual interior. Therefore, to enable RDAL to jointly train the layout and residual depth of the indoor scene, we design $G_{out}$ to have three decoding branches.
Each of them generates RGB panorama $I^{rgb}_{out}$, layout depth panorama $I^{d,lay}_{out}$, and residual depth panorama $I^{d,res}_{out}$ respectively, as shown in Fig.~\ref{fig:generator}. 
Intuitively, $I^{d,lay}_{out}$ determines the layout structure and $I^{d,res}_{out}$ determines the interior objects. 
Then, element-wise addition of $I^{d,lay}_{out}$ and $I^{d,res}_{out}$ gives the output total depth map panorama. 

\vspace{-12pt}
\subsubsection{Discriminator}
\label{disc}
\vspace{-5pt}
We use the multi-scale discriminator $D$ from \cite{wang2018high}, but modify it to have five input and output channels (three for $I^{rgb}$, one for $I^{d,lay}$, one for $I^{d,res}$). The detailed discriminator structure can be found in the suppl. material.

\vspace{-10pt}
\subsubsection{Loss Function}
\vspace{-5pt}

For training $G$, we use weighted sum of pixel-wise L1 loss and adversarial loss. The pixel-wise L1 loss between the GT and the output panorama, denoted as $L_{pixel}$, consists of three terms as the $G$ has three outputs (RGB, layout depth, residual depth panorama):
\begin{equation}
    L^{total}_{pixel} = L^{rgb}_{pixel} + L^{d,lay}_{pixel} + L^{d,res}_{pixel}.
\end{equation}

For the adversarial loss $L_{adv}$, we used LSGAN loss~\cite{mao2017least}: $L_{adv} =  \frac{1}{2}\mathop{\mathbb{E}}{ [   (D({I^{total}_{out}})-1)^2] }$, 
where $I^{total}_{out}$ is concatenation of generator outputs $I^{rgb}_{out}$, $I^{d,lay}_{out}$ and $I^{d,res}_{out}$, and $D$ is a discriminator trained to output $one$ for GT and $zero$ for $I^{total}_{out}$ with MSE loss. 
By decomposing the total depth loss into $L^{d,lay}$ and $L^{d,res}$, our RDAL scheme allows the generator $G$ to synthesize RGB-D panorama that generates highly plausible interior.
Finally, the total loss for generator is: 
\begin{equation}
    L_G=\lambda L^{total}_{pixel} + L_{GAN}\\[-5pt]
\end{equation}
where $\lambda$ is a weighting factor. The generator $G$ is trained by minimizing the total loss $L_G$. 
Detailed loss terms can be found in the suppl. material.



\subsection{Fréchet Auto-Encoder Distance (FAED)}
\label{subsec:FAED}

\vspace{-0.15cm}
\subsubsection{Auto-Encoder Network} 
\vspace{-0.1cm}
\label{subsubsec:A}

\begin{figure}[t]
\begin{center}
   \hspace*{-0.0\linewidth}\includegraphics[width=.91\linewidth]{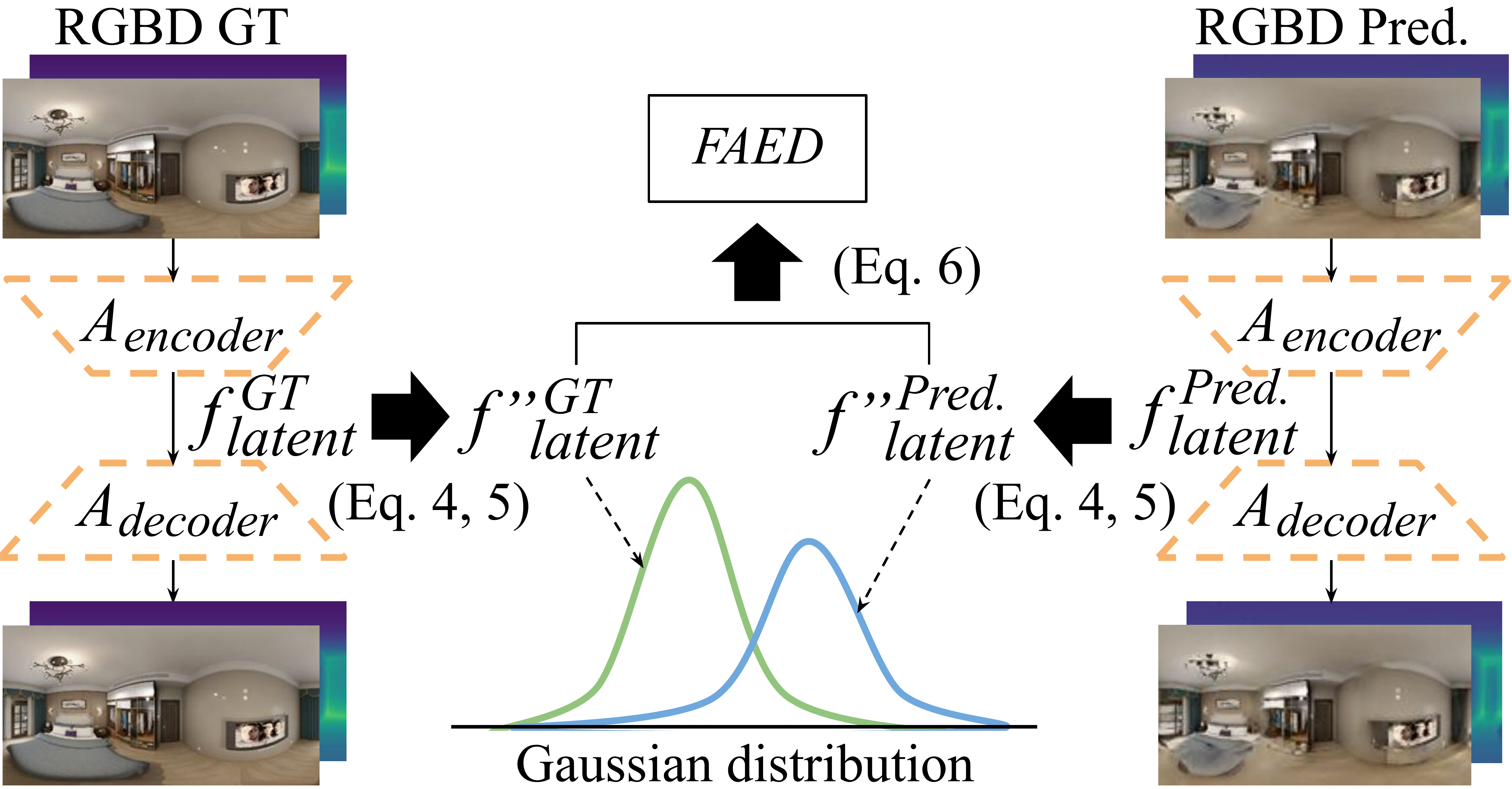}
\end{center}
   \vspace{-0.5cm}
   \caption{The proposed FAED metric for RGB-D panorama quality evaluation. It measures the distance of the distributions of latent features extracted from the pre-trained auto-encoder network on RGB-D panorama.
}
\label{fig:autoencoder}
\vspace{-15pt}
\end{figure}

Similar to the high-level features in a CNN trained with large-scale semantic labels, latent features $f_{latent}$ in a trained auto-encoder also contain high-level information, as it is forced to reconstruct the input from the latent features. Therefore, we propose to train  an  auto-encoder, which can be done without any labels in the dataset, and use the latent features in the auto-encoder to extract perceptually meaningful information. In this way, performance evaluation can be performed for any data that lacks a labeled dataset.
The auto-encoder $A$ consists of an encoder and decoder: $A_{encoder}$ and $A_{decoder}$, as shown in Fig.~\ref{fig:autoencoder}. 
The detailed structure of $A$ is given in the suppl. material.


\vspace{-0.25cm}
\subsubsection{Calculation of FAED for RGB-D Panorama} 
\vspace{-0.1cm}
We denote 
$f_{latent}$ at $c$-th channel, $h$-th row, and $w$-th column as $f_{latent}(c,h,w)$.
Note that as we use ERP, the $h$ and $w$ has one-to-one relation to latitude and longitude. 


\noindent\textbf{Longitudinal Invariance} To evaluate the performance of $G$, we extract $f_{latent}$ from generated samples using $A_{encoder}$. However, as we generate the upright ERP image, it is expected to have a distance metric that is invariant to the longitudinal shift. This is because an upright ERP panorama represents the same scene when it's cyclically shifted in the longitudinal direction. Therefore, to make the resulting distance metric invariant to the longitudinal shift, we take the mean for the longitudinal direction of $f_{latent}$ as:
\vspace{-4pt}
\begin{equation}
    f_{latent}^\prime(c,h) = \frac{1}{W} \sum_{w} f_{latent}(c,h,w). \\[-4pt]
\end{equation}

\noindent\textbf{Latitudinal Equivariance} 
As the ERP has varying sampling rates depending on the latitude $\phi$, we apply different weights on $f_{latent}^\prime$ based on the latitude.
Specifically, we multiply $cos(\phi)$ to feature at the latitude $\phi$, because in ERP, each pixel occupies $cos(\phi)$ area in the spherical surface, compared with the pixels in the equator. Formally, the resulting feature $f_{latent}^{\prime\prime}$ is expressed as:
\vspace{-4pt}
\begin{equation}
    f_{latent}^{\prime\prime}(c,h) = \cos \phi \cdot f_{latent}^\prime(c,h).
\end{equation}

\begin{table*}[t!]

\caption{Quantitative results of RGB panorama synthesis on Structured3D dataset. 
As \cite{sumantri2020360} uses 4 identical perspective RGB masks on horizontal central line, we report our results in same setting. In other cases, we follow designed arbitrary configuration of RGB sensor that uses 1\~4 number of inputs.
Zero number of depth input means that depth map is not used for RGB panorama synthesis. 
For FAED calculation, GT depth is used along with synthesized RGB panorama.  \textbf{Bold} numbers indicate the best results.
}
\label{tab:result_rgb}
\vspace{-8pt}
\footnotesize
\resizebox{\textwidth}{!}{%
\begin{tabular}{c|c|cc|ccc|c|c}
\hline
\multirow{2}{*}{Category}   & \multirow{2}{*}{Method} & \multicolumn{2}{c||}{Input no. ($n$)}     & \multicolumn{3}{c|}{RGB metric}                                                               & Layout metric            & Proposed metric  \\ \cline{3-9} 
                            &                         & \multicolumn{1}{c|}{RGB}   & \multicolumn{1}{c||}{Depth} & \multicolumn{1}{c|}{PSNR($\uparrow$)} & \multicolumn{1}{c|}{SSIM($\uparrow$)} & LPIPS($\downarrow$) & 2D Corner error($\downarrow$) & FAED($\downarrow$) \\ \Xhline{3\arrayrulewidth}
\multirow{3}{*}{Inpainting} & BRGM~\cite{marinescu2020bayesian}                    & \multicolumn{1}{c|}{\multirow{5}{*}{1/2/3/4}} & \multicolumn{1}{c||}{\multirow{5}{*}{0}}     & \multicolumn{1}{c|}{14.00}               & \multicolumn{1}{c|}{0.5310}               & 0.6192                  & 72.52                    & 442.3            \\ \cline{2-2} \cline{5-9}  
                            & CoModGAN~\cite{zhao2021large}                & \multicolumn{1}{c|}{} & \multicolumn{1}{c||}{}     & \multicolumn{1}{c|}{14.35}          & \multicolumn{1}{c|}{0.5837}         & 0.4768            & 62.45                    & 208.2            \\ \cline{2-2} \cline{5-9} 
                            & LaMa~\cite{suvorov2021resolution}                    & \multicolumn{1}{c|}{} & \multicolumn{1}{c||}{}     & \multicolumn{1}{c|}{13.74}          & \multicolumn{1}{c|}{0.5207}         & 0.5658            & 51.12                    & 379.2            \\ \cline{1-2} \cline{5-9}
Outpainting                 & Boundless~\cite{boundless}               & \multicolumn{1}{c|}{} & \multicolumn{1}{c||}{}     & \multicolumn{1}{c|}{13.74}          & \multicolumn{1}{c|}{0.5663}         & 0.6144            & 74.47                    & 429.4            \\ \cline{1-2} \cline{5-9}
\multirow{1}{*}{}   & Ours                    & \multicolumn{1}{c|}{} & \multicolumn{1}{c||}{}     & \multicolumn{1}{c|}{\textbf{16.21}}          & \multicolumn{1}{c|}{\textbf{0.6161}}         &\textbf{ 0.4549 }           & \textbf{39.63 }                   & \textbf{162.3}            \\
\Xhline{3\arrayrulewidth}
Panorama syn.              & Sumantri \etal \cite{sumantri2020360}& \multicolumn{1}{c|}{\multirow{2}{*}{4}}     & \multicolumn{1}{c||}{\multirow{2}{*}{0}}     & \multicolumn{1}{c|}{\textbf{18.49}}          & \multicolumn{1}{c|}{\textbf{0.6680}}         & 0.4190            & 50.76                    & 443.4            \\ \cline{1-2} \cline{5-9}
                    & Ours                    & \multicolumn{1}{c|}{}     & \multicolumn{1}{c||}{}     & \multicolumn{1}{c|}{17.29}               & \multicolumn{1}{c|}{0.6510}               & \textbf{0.3975}                  & \textbf{34.68}                         & \textbf{103.1}                 \\ \hline
                   
\end{tabular}%
}
\vspace{-6pt}
\end{table*}

\begin{figure*}[t!]
\begin{center}

  \hspace*{-0.0\linewidth}\includegraphics[width=\textwidth]{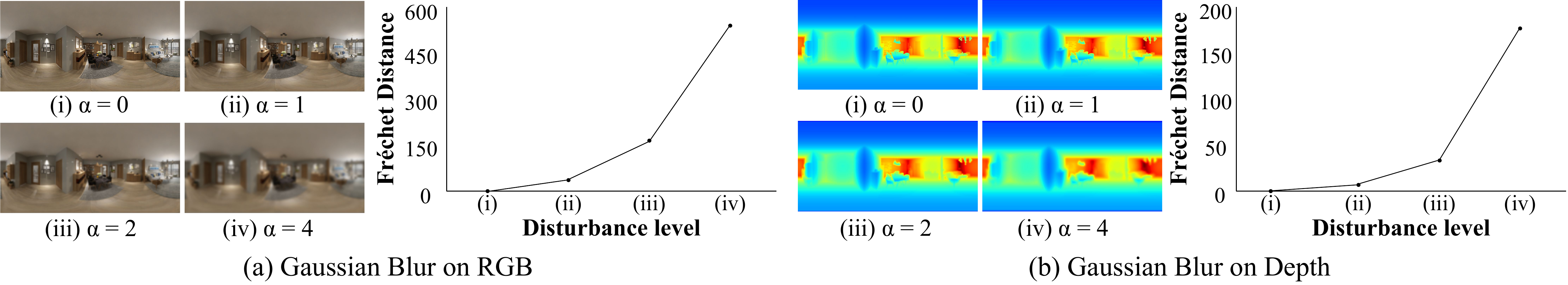}
\end{center}
  \vspace{-0.7cm}
  \captionsetup{font=small}
  \caption{Verification of FAED in Structured3D dataset. It can be seen that FAED correlates well perceptual evaluation of human, as FAED increases as the data becomes more corrupted. For more detailed results, please refer to the suppl. material.}
  \vspace{-0.6cm}
\label{fig:verification}
\end{figure*}


\noindent{}\textbf{Fréchet Distance} We treat the resulting $f_{latent}^{\prime\prime}$ as a vector and assume that it has a multi-dimensional Gaussian distribution. Then, we get the distribution of ground truths $\mathcal{N}(m, C)$ and that of generated samples $\mathcal{N}(\hat{m},\hat{C})$, and calculate the Fréchet distance $d$ between them as given by \cite{dowson1982frechet}: 
\begin{equation}
\begin{split}
    d^2(\mathcal{N}(m,&C),\mathcal{N}(\hat{m},\hat{C}))\\
    &= ||m-\hat{m}||_2^2
    +Tr(C+\hat{C}-2(C\hat{C})^{1/2}). \\[-4pt]
\end{split}
\end{equation}
We use $d^2$ as a perceptual distance metric where $m$ and $C$ denote mean and covariance, respectively.


\vspace{-4pt}
\section{Experimental Results}
\vspace{-3pt}
\noindent\textbf{Synthetic Dataset}
\label{subsec:dataset}
Structured3D dataset~\cite{zheng2019structured3d} provides various textures of indoor scenes with a $512\times 1024$ resolution. 
We split the dataset into train, validation, and test set
where the numbers of data are 17468, 2183, and 2184, respectively. 
In addition, with the corner locations 
provided in the dataset, we manually generated layout depth maps of each 3D scene. 
The residual depth maps are obtained by subtracting the layout depth from the GT depth map.    

\noindent\textbf{Real Dataset}
We used a combination of two datasets: Matterport3D \cite{Matterport3D}, and 2D-3D-S dataset \cite{armeni2017joint}. Both datasets provide real-world indoor RGB-D panorama. Since this dataset does not provide sufficient number of annotated layout, it can't be used for training our framework and only used for test purpose.
We excluded data with too many of invalid pixels from test dataset, then its number of data is 603. 


\noindent\textbf{Implementation Details}
For the details about our implementation, please refer to the supplementary material.
\vspace{-5pt}
\subsection{Verification of FAED}
\label{subsec:verification_faed}
\vspace{-3pt}
To show the effectiveness of FAED on measuring the perceptual quality of RGB-D panorama, we corrupt the Structured3D dataset \cite{zheng2019structured3d} in two ways: corrupting RGB images only and corrupting depth maps only. 
Following \cite{heusel2017gans}, we corrupt the dataset by applying various types of noise: Gaussian blur, Gaussian noise, uniform patches, swirl, and salt and pepper noise. 
Here, we only show the plots for Gaussian blur in Fig.~\ref{fig:verification} due to the lack of space. Other results can be found in suppl. material.
\textit{Note that the evaluation is done for RGB-D panorama, neither for RGB image alone nor for depth map alone.}
As shown in Fig. \ref{fig:verification}, the Fréchet distance for both RGB and depth panorama increases as the disturbance level (Gaussian blur) is increased. 
We show that the same applies to the other four types of noises in the supplementary material.
\textit{This indicates the perceptual quality of RGB-D panorama becomes poorer as the FAED score increases.}


\begin{figure}[t]
\begin{center}
  \hspace*{-0.0\linewidth}\includegraphics[width=\linewidth]{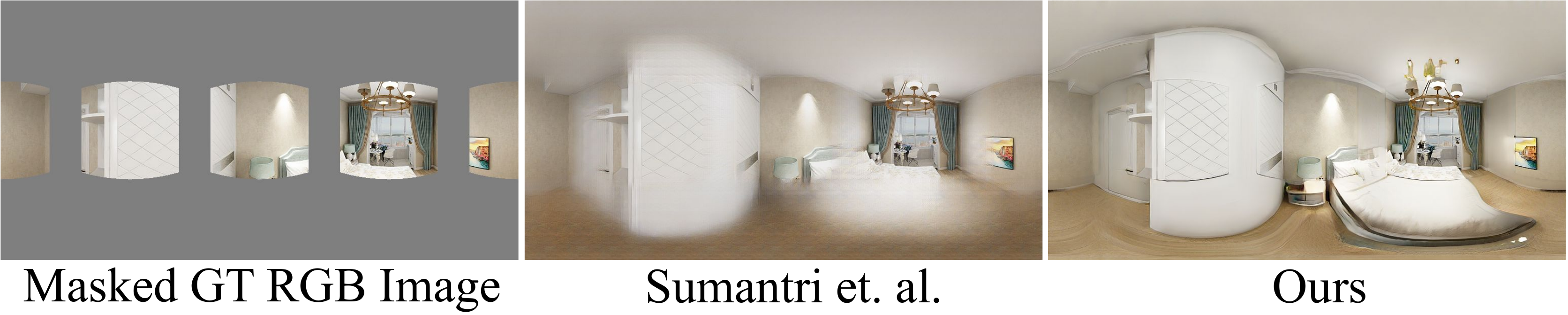}
\end{center}
  \vspace{-0.7cm}
  \caption{Qualitative comparison to Sumantri et. al.~\cite{sumantri2020360}. While the result from~\cite{sumantri2020360} is blurry, our result is sharp and realistic.}
\label{fig:pano_syn_res}
\vspace{-15pt}
\end{figure}

\begin{table*}[h!]
\caption{Quantitative results of depth panorama synthesis on Structured3D dataset. Depth input type L/P means that we use LiDAR (L) and dense perspective depth sensor (P) in arbitrary configurations for the input. Full RGB image is used along with synthesized depth panorama for FAED calculation. \textbf{Bold} numbers indicate the best results.
}
\label{tab:results_depth}
\vspace{-8pt}
\tiny
\resizebox{\textwidth}{!}{%
\begin{tabular}{c|c|cc|cc|c|c}
\hline
\multirow{2}{*}{Category}         & \multirow{2}{*}{Method} & \multicolumn{2}{c||}{Input type}        & \multicolumn{2}{c|}{Depth metric}  & Layout metric & Proposed metric \\ \cline{3-8} 
                                  &                         & \multicolumn{1}{c|}{RGB}   & \multicolumn{1}{c||}{Depth} & \multicolumn{1}{c|}{AbsREL($\downarrow$)} & RMSE($\downarrow$) & 2D IoU($\uparrow$)        & FAED($\downarrow$)            \\ \Xhline{3\arrayrulewidth}
\multirow{4}{*}{Depth syn.} & CSPN      \cite{Cheng_2020_TPAMI}              & \multicolumn{1}{c|}{\multirow{5}{*}{Full}}  & \multicolumn{1}{c||}{\multirow{5}{*}{L/P}}   & \multicolumn{1}{c|}{0.0855}      & 2214    & 0.8062             & 428.9               \\ \cline{2-2} \cline{5-8} 
                                  & NLSPN  \cite{park2020non}                 & \multicolumn{1}{c|}{}  & \multicolumn{1}{c||}{}   & \multicolumn{1}{c|}{0.1268}      & 2807    & 0.7333             & 836.1               \\ \cline{2-2} \cline{5-8} 
                                  & MSG-CHN  \cite{Li_2020_WACV}               & \multicolumn{1}{c|}{}  & \multicolumn{1}{c||}{}   & \multicolumn{1}{c|}{0.1764}      & 3296    & 0.6724             & 896.4               \\ \cline{2-2} \cline{5-8}
                                  & PENet      \cite{hu2020PENet}             & \multicolumn{1}{c|}{}  & \multicolumn{1}{c||}{}   & \multicolumn{1}{c|}{0.1740}      & 3145    & 0.7033             & 906.0               \\ \cline{1-2} \cline{5-8}
         & Ours                     & \multicolumn{1}{c|}{}  & \multicolumn{1}{c||}{}   & \multicolumn{1}{c|}{\textbf{0.0844}}      & \textbf{1942}    & \textbf{0.8286}             & \textbf{131.5}                        \\ \hline
\end{tabular}%
}
\end{table*}

\begin{figure*}[t!]
\begin{center}

  \hspace*{-0.0\linewidth}\includegraphics[width=\textwidth]{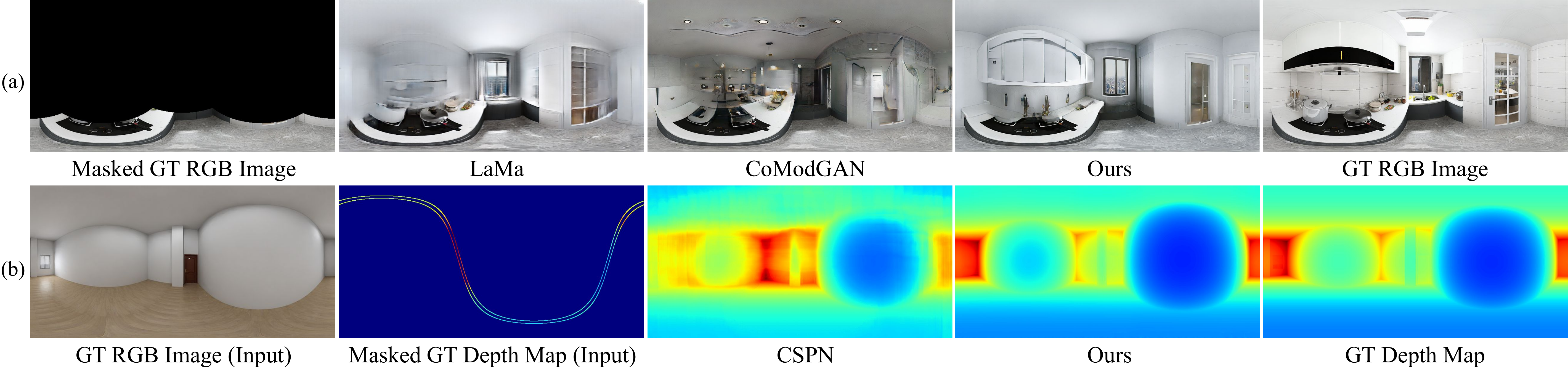}
\end{center}
  \vspace{-0.6cm}
  \captionsetup{font=small}
  \caption{(a) Visual results for RGB panorama synthesis on Structured3D dataset. Two methods, LaMa and CoMoGAN,
  are visualized for comparison. (b) Visual results for depth panorama synthesis on Structured3D dataset. 
  CSPN is also visualized for comparison. More qualitative results can be found in suppl. material.
  }
  \vspace{-0.6cm}
\label{fig:inpainting_res}
\end{figure*}

\vspace{-2pt}
\subsection{RGB-D Panorama Synthesis}
\vspace{-4pt}

\begin{table}[t!]
\caption{Ablation study results of BIPS framework. The experiments in four rows take RGB-D input.
}
\label{tab:ablation}
\vspace{-8pt}
\footnotesize
\renewcommand{\tabcolsep}{17pt}
\begin{tabular}{c|cc}
\hline
\multirow{2}{*}{Method} & \multicolumn{2}{c}{Metric}         \\ \cline{2-3} 
                        & \multicolumn{1}{c|}{2D IoU($\uparrow$)} & FAED($\downarrow$)  \\ \Xhline{3\arrayrulewidth}
IwDS (\cite{zhao2021large} + \cite{Cheng_2020_TPAMI})                  & \multicolumn{1}{c|}{0.7561} & 640.9 \\ \hline
Ours w/o BFF            & \multicolumn{1}{c|}{0.7859} & 381.4 \\ \hline
Ours w/o RDAL           & \multicolumn{1}{c|}{0.7164} & 329.0 \\ \hline
Ours                    & \multicolumn{1}{c|}{\textbf{0.8158}} & \textbf{198.0} \\ \hline
\end{tabular}
\vspace{-13pt}
\end{table}

\begin{figure*}[t]
\begin{center}
   \hspace*{-0.0\linewidth}\includegraphics[width=\linewidth]{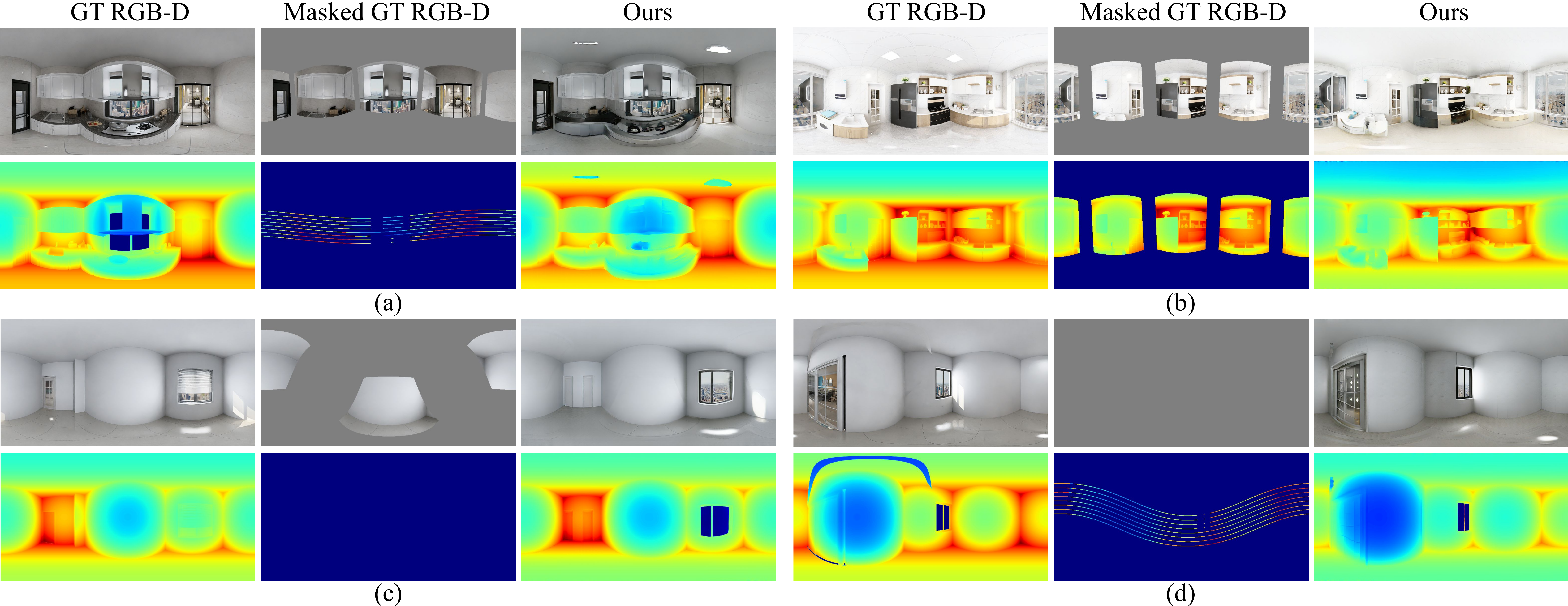}
\end{center}
   \vspace{-0.3cm}
   \vspace{-8pt}
   \caption{
Visualization of our synthesized RGB-D panorama results using RGB-D data in arbitrary configurations. (a) and (b) take both RGB and depth data, (c) takes only RGB and (d) takes only depth data. More results are visualized in suppl. materials.}
\label{fig:our_syn_res}
\vspace{-8pt}
\end{figure*}

\begin{figure*}[t]
\begin{center}
   \hspace*{-0.0\linewidth}\includegraphics[width=\linewidth]{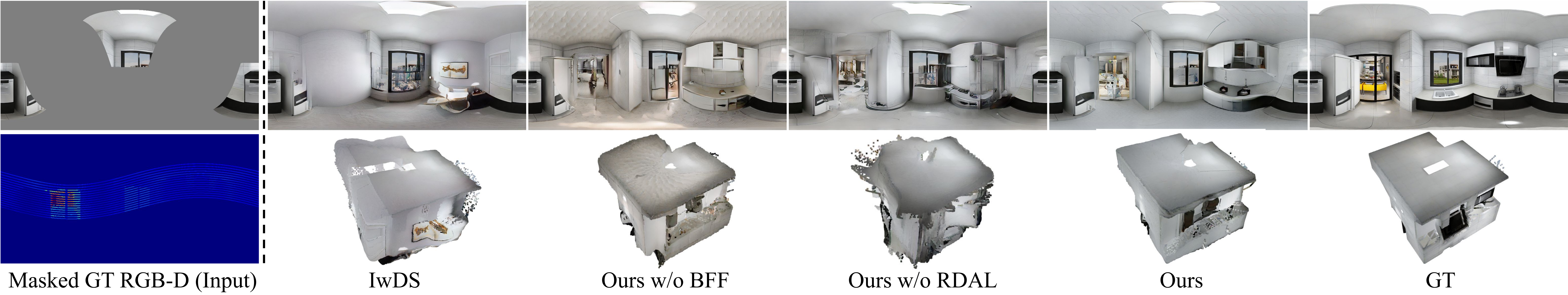}
\end{center}
   \vspace{-0.3cm}
   \vspace{-8pt}
   \caption{
Visualization of ablation study results on Structured3D dataset. Result without BFF shows artifacts irrelevant to given input and result without RDAL infers distorted room layout while our final model synthesizes perceptual undistorted indoor room.
   }
\label{fig:ablation}
\vspace{-15pt}
\end{figure*}

\begin{figure}[t]
\begin{center}
  \hspace*{-0.0\linewidth}\includegraphics[width=0.98\linewidth]{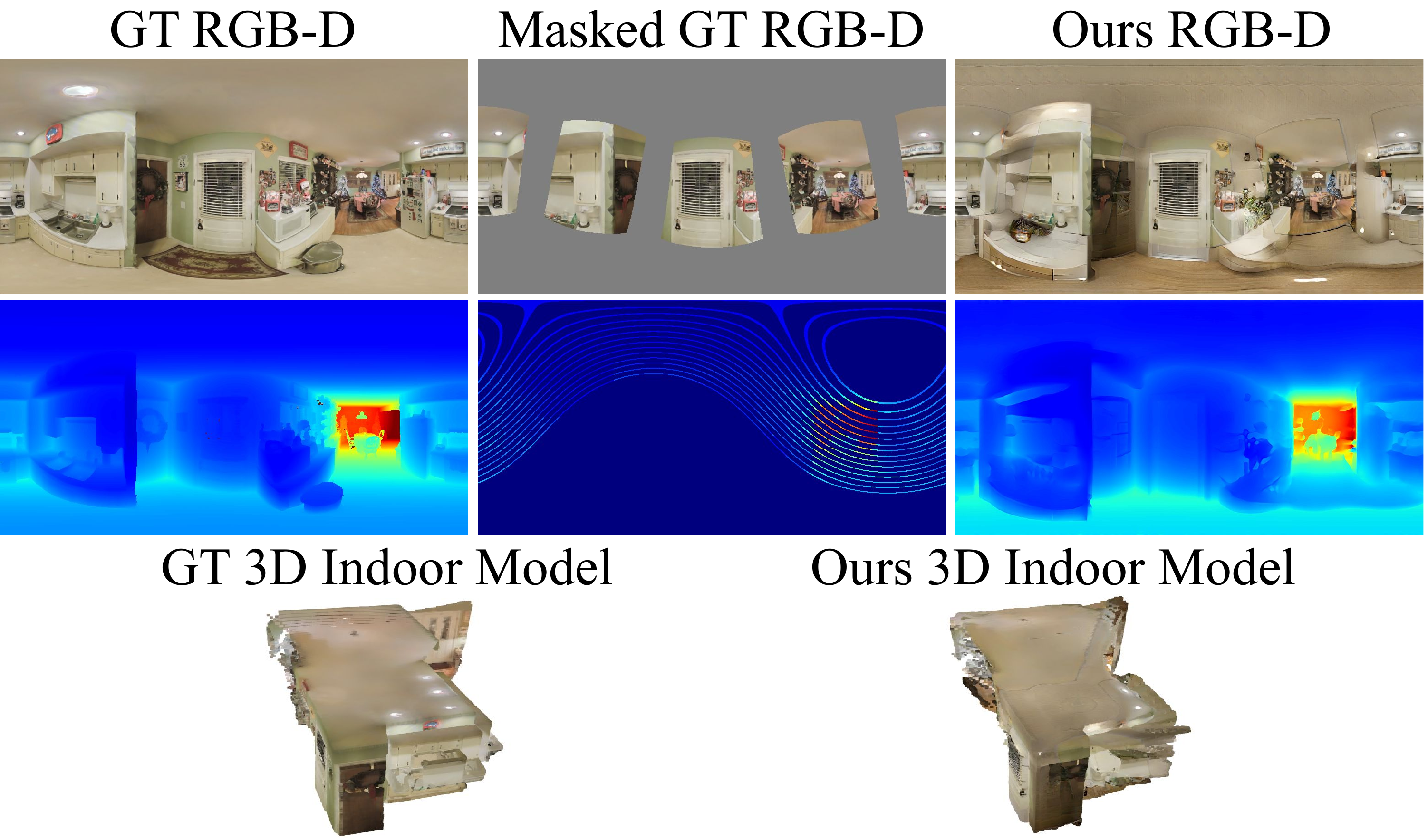}
\end{center}
  \vspace{-0.5cm}
  \caption{Visualization of our synthesized RGB-D panorama and 3D indoor model on Matterport3D dataset.}
\label{fig:matter}
\vspace{-18pt}
\end{figure}

\noindent\textbf{RGB Panorama Evaluation}
Table~\ref{tab:result_rgb} shows the quantitative comparison with the inpainting and outpainting methods on the Structured3D dataset. We use PSNR, SSIM and LPIPS to evaluate the quality of RGB panorama. 
We also measure 2D corner error, where the 2D GT corner points are compared with the estimated 2D corner points using DuLa-Net~\cite{yang2019dula} on the synthesized RGB panorama.
We also use the proposed FAED  to jointly evaluate the quality of RGB-D information. 

As shown in Table~\ref{tab:result_rgb}, our method outperforms the image inpainting and outpainting methods: BRGM~\cite{marinescu2020bayesian}, CoModGAN~\cite{zhao2021large}, LaMa~\cite{suvorov2021resolution} and Boundless~\cite{boundless}, by a large margin for all metrics. For instance, our method outperforms the best inpainting method, CoModGAN, by an 4.6\% decrease of LPIPS score, 36.5\% drop of 2D corner error, and 22\% decline of FAED score.  The effectiveness can also be visually verified in Fig.~\ref{fig:inpainting_res}(a).
Our method produces clearer RGB panorama images compared with LaMa producing blurry images. Although CoModGAN produces clear RGB outputs, it doesn't consider the indoor layout and semantic information of the furniture, \eg electric cooker is combined with bookshelves, as shown in Fig.~\ref{fig:inpainting_res}.
Therefore, its layout is semantically inconsistent with the input RGB region and its FAED score is higher than ours.

We also compare with the panorama synthesis method, Sumantri \etal~\cite{sumantri2020360}. 
Our method shows slightly lower scores using the conventional metrics, PSNR and SSIM; however, it shows the much better LPIPS score (0.3975 vs 0.4190), 2D corner error (34.68 vs 50.76) and FAED score (103.1 vs 443.4), respectively. We argue that PSNR and SSIM merely measure local photometric similarity, and thus fail to well reflect the perceptual quality.
This can be verified from Fig.~\ref{fig:pano_syn_res} where we visually compare with \cite{sumantri2020360}. Our method synthesizes better textures and shows much higher visual quality. More results can be found in suppl. material.  




\noindent\textbf{Depth Panorama Evaluation} 
We compare our method with the image-guided depth synthesis methods on Structured3D dataset. 
To evaluate the quality, we use AbsREL and RMSE, and the proposed FAED. We also use layout 2D IoU, as was done in~\cite{choi20203d}.
The details of the metrics and results can be found in the supplementary material. 

Table~\ref{tab:results_depth} shows the quantitative comparison with the depth synthesis methods: CSPN~\cite{Cheng_2020_TPAMI}, NLSPN~\cite{park2020non}, MSG-CHN~\cite{Li_2020_WACV} and PENet~\cite{hu2020PENet}. In particular, our method outperforms on of the best depth synthesis method, CSPN, with much better AbsREL score (0.0844 vs 0.0855), RMSE (1942 vs 2214), 2D IoU (0.8286 vs 0.8062) and FAED score (131.5 vs 428.9).
With the proposed RDAL scheme, our method estimates the best layout depth, which is demonstrated by the highest layout 2D IoU. 
This in turn, considerably affects the overall depth error in other metrics as well. 
Figure.~\ref{fig:inpainting_res}(b) shows the qualitative comparison with CSPN~\cite{Cheng_2020_TPAMI}. 
CSPN synthesizes the interior components,~\eg, beds, relatively well; however, the depth of planes (\eg, walls and ceiling) are not clear. Therefore, it cannot synthesize a valid layout,  
failing to generate a realistic 3D indoor model. 
By contrast, our synthesized depth panorama shows an undisturbed and clear layout. 


\noindent{}\textbf{Evaluation on Real Dataset}
We evaluated our synthesized RGB-D panorama on real indoor scenes in Matterport3D and 2D-3D-S dataset. An output RGB-D panorama and its 3D indoor model are visualized in Fig.~\ref{fig:matter}. 
Overall, our method synthesizes high-quality RGB-D panorama on real indoor scenes, unseen during training. Our synthesized depth panorama shows precise indoor layout and plausible residuals, generating a realistic 3D indoor model. For quantitative result, our method achieved better FAED score than IwDS (4645 vs 5099). Since the FAED score between synthetic and real dataset is 3517, it demonstrates that there exists distribution shift between the datasets and our result shows realistic RGB-D panorama and 3D indoor models consistently in both synthetic and real indoor scenes.
More results can be found in suppl. material.

\vspace{-8pt}
\subsection{Ablation Study and Analysis}
\vspace{-3pt}
\noindent\textbf{Inpainting w/ Depth Synthesis (IwDS)} One solution to obtain RGB-D panorama from partial RGB-D inputs is sequentially accomplishing RGB synthesis (inpainting) and depth synthesis. 
To be specific, a RGB panorama is first synthesized from partial RGB input using the image inpainting method. 
Then, depth panorama is synthesized by applying the depth synthesis method to the synthesized RGB panorama and partial depth input.
We chose CoModGAN \cite{zhao2021large} and CSPN \cite{Cheng_2020_TPAMI} for RGB and depth synthesis methods, which showed the highest FAED score in Table~\ref{tab:result_rgb} and Table~\ref{tab:results_depth}.
In Table~\ref{tab:ablation}, it can be seen that IwDS leads lower 2D IoU score and a much higher FAED score than our method. 
This indicates that the two-stage, sequential synthesis of RGB-D panorama is less effective than our BIPS framework that fuses the bi-modal features, trained with one-stage, joint learning scheme. Also, IwDS fails to generate realistic 3D indoor models, with distorted indoor layouts and severe bumpy surfaces as shown in Fig.~\ref{fig:ablation}.

\noindent\textbf{Impact of BFF} We study the effectiveness of RGB-D panorama synthesis by removing the BFF branch in the generator. 
In details, $G_{BFF}$ is replaced with a single branch network taking the concatenation of $G_{in}^{rgb}(I_{in}^{rgb})$ and $G_{in}^{depth}(I_{in}^{d})$. 
As shown in Table~\ref{tab:ablation}, the 2D IoU drops and FAED score increases without BFF.  Fig.~\ref{fig:ablation} shows that texture of the RGB-D output is not consistent with the given arbitrary RGB-D input. This reflects that BFF significantly contributes to well-process the bi-modal information.

\noindent\textbf{Impact of RDAL} We further validate the effectiveness of RDAL by comparing the results without RDAL.
The number of output branches are reduced to two, and each are designed to learn RGB and total depth panorama, respectively. 
As shown in Table~\ref{tab:ablation}, the 2D IOU score drops, and FAED score increases without RDAL. It shows that RDAL is critical for estimating precise indoor layout. The impact of RDAL is visually verified in Fig.~\ref{fig:ablation}. The result without RDAL shows distorted indoor layout while having fewer artifacts than ours w/o BFF. In summary, dividing the total depth into layout and residual depth helps to synthesize more structural 3D indoor model. 

\vspace{-8pt}
\section{Conclusion}
\vspace{-3pt}
In this paper, we tackled a novel problem of synthesizing RGB-D indoor panoramas from arbitrary configurations of RGB and depth 
inputs. Our method can synthesize high-quality RGB-D panoramas with the proposed BIPS framework by utilizing the bi-modal information and jointly training the layout and residual depth of indoor scenes. 
Moreover, 
a novel evaluation metric FAED was proposed and its validity was demonstrated. Extensive experiments show that our method achieves the SoTA RGB-D panorama synthesis performance. 

\noindent \textbf{Limitation} 
We mainly focused on 
indoor scenes, and proposed RDAL is hardly applicable to outdoor scenes. 
Future work will extend our method to various environments.

\clearpage
{\small
\bibliographystyle{ieee_fullname}
\bibliography{egbib}
}

\end{document}